\PassOptionsToPackage{svgnames}{xcolor}
\documentclass[sigconf]{acmart}

\usepackage{booktabs} 

\usepackage{booktabs} 
\usepackage{url,paralist,multirow,booktabs}
\usepackage{color}

\usepackage{amsfonts}
\usepackage{bm}
\usepackage{amsmath}
\usepackage{pifont}
\newcommand{\cmark}{\ding{51}}%
\newcommand{\xmark}{\ding{55}}%

\copyrightyear{2017} 
\acmYear{2017} 
\setcopyright{acmcopyright}
\acmConference{CIKM'17 }{November 6--10, 2017}{Singapore, Singapore}\acmPrice{15.00}\acmDOI{10.1145/3132847.3133110}
\acmISBN{978-1-4503-4918-5/17/11}

\begin{document}
\title[Hierarchical RNN for Text-Based Speaker Change Detection]{Hierarchical RNN with Static Sentence-Level Attention\\
	for Text-Based Speaker Change Detection}

\author{Zhao Meng}\affiliation{%
  \institution{Key Laboratory of High Confidence Software Technologies, MoE\\
  	Software Institute, Peking University\\
}
}
\email{zhaomeng.pku@outlook.com}

\author{Lili Mou}
\affiliation{%
  \institution{AdeptMind Research\\Toronto, Canada}
}
\email{doublepower.mou@gmail.com
}
\email{
lili@adeptmind.ai}

\author{Zhi Jin}
\authornote{Corresponding author. This research is supported by the National Basic Research Program of China (the 973 Program) under Grant No.~2015CB352201, and the National Natural Science Foundation of China under Grant Nos.~61232015 and 61620106007.}
\affiliation{%
  \institution{Key Laboratory of High Confidence Software Technologies, MoE\\
  	Software Institute, Peking University}
}
\email{zhijin@sei.pku.edu.cn}

\renewcommand{\shortauthors}{Meng, Mou, and Jin}

\begin{abstract}
This paper provides a sample of a \LaTeX\ document which conforms,
somewhat loosely, to the formatting guidelines for
ACM SIG Proceedings.\footnote{This is an abstract footnote}
\end{abstract}

%
%

\begin{abstract}
	Speaker change detection (SCD) is an important task in dialog modeling.
	Our paper addresses the problem of text-based SCD, which differs from existing audio-based studies and is useful in various scenarios, for example, processing dialog transcripts where speaker identities are missing (e.g., OpenSubtitle), and enhancing audio SCD with textual information.
	We formulate text-based SCD as a matching problem of utterances before and after a certain decision point; we propose
	a hierarchical recurrent neural network (RNN) with static sentence-level attention.
	Experimental results show that neural networks consistently achieve better performance than feature-based approaches, and that
	our attention-based model significantly outperforms non-attention neural networks.\footnote{Code available at \url{https://sites.google.com/site/textscd/}}
\end{abstract}

\maketitle

\section{Introduction}\label{sec:introduction}

Speaker change detection (SCD), or sometimes known as \textit{speaker segmentation}, aims to find changing points of speakers in a dialog. Specifically, a speaker change occurs when the current and the next sentences are not uttered by the same speaker~\cite{Anguera2012Speaker}. 
Detecting speaker changes plays an important role in dialog processing, and is a premise of dialog understanding, speaker clustering~\cite{sinha2005cambridge}, etc.

In this paper, we address the problem of text-based speaker change detection, which differs from traditional SCD with audio input~\cite{lu2005unsupervised,kemp2000strategies}. Text-based SCD is important for several reasons:

$\bullet$ Evidence in the speech processing domain shows that text information can improve speech-based SCD~\cite{li2009improving}. However, there lacks specialized research for text-based SCD.

$\bullet$ In some scenarios, researchers may not have access to raw audio signals for SCD. 
Vinyals et al.~\cite{vinyals2015neural} and Li et al.~\cite{li2015diversity}, for example,  train sequence-to-sequence neural networks to automatically generate replies in an open-domain dialog system. They use OpenSubtitle~\cite{tiedemann2009news} as the corpus, but assume every two consecutive sentences are uttered by different speakers (which brings much noise to their training data).

$\bullet$ The fast development of dialog analysis puts high demands on understanding textual data~\cite{vinyals2015neural,cikm2, yancikm1}---in addition to audio features alone---because human-computer conversation involves deep semantics, requiring complicated natural language processing. Text-based SCD could also serve as a surrogate task for general speaker modeling, similar to next utterance classification (NUC) being a surrogate task for general dialog generation~\cite{ubuntu}.

Using only text to detect speaker changes brings new challenges. Previous audio-based SCD depends largely on acoustic features, e.g., pitch~\cite{lu2005unsupervised}  and silence points~\cite{kemp2000strategies}, which provide much information of speaker changes. With textual features alone, we need deeper semantic understanding of natural language utterances. 

\begin{figure}[!t]
	\centering
	\includegraphics[width=1.05\linewidth]{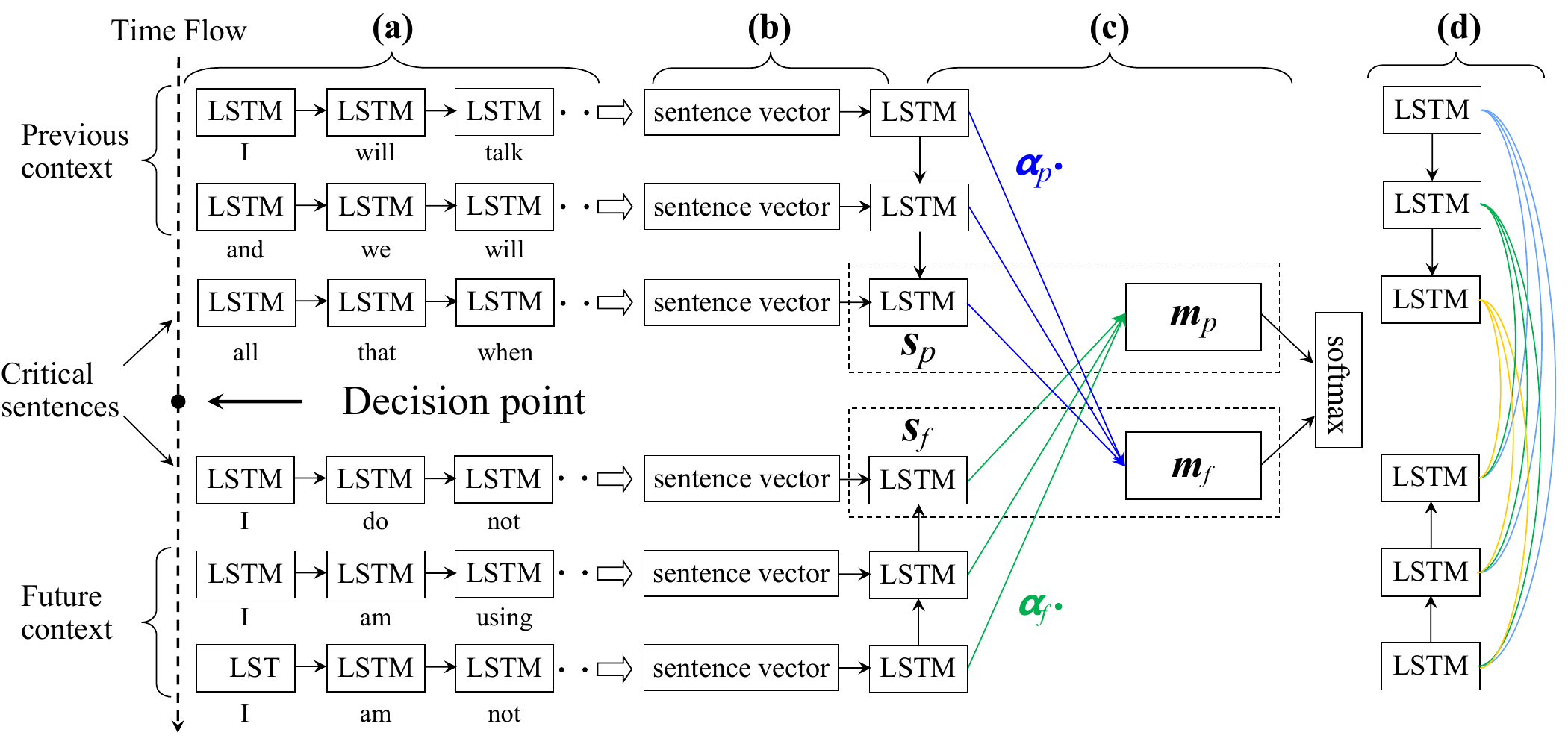}
	\caption{The proposed neural network. (a) LSTM-RNN sentence encoder. (b) Context encoder with another LSTM-RNN. (c) Sentence-level static attention. (d) C.f.~Dynamic attention.
		Notice that we consider text-based speaker change detection in this paper, so we have future utterances as context.}
	\label{fig:model}
	
\end{figure}

In this paper, we formulate text-based SCD as a binary sentence-pair classification problem, that is, we would like to judge whether the speaker is changing between each consecutive sentence pair (which we call a \textit{decision point}). We also take into consideration previous and future sentences around the current decision point as context (Figure~\ref{fig:model}), serving as additional evidence.\footnote{Because our task is based on text (and is \textit{not} online speaker change detection), we actually have access to the ``future context'' after the decision point.}

We propose a hierarchical RNN with static sentence-level attention for text-based speaker change detection. First, we use a long short term memory (LSTM)-based recurrent neural network (RNN) to capture the meaning of each sentence. Another LSTM-RNN integrates sentence information into a vector, before and after the decision point, respectively; the two vectors are combined for prediction. To better explore the context, we further apply an attention mechanism over sentences to focus on relevant information during context integration.  Compared with widely-used word-level attention, our sentence-level attention is more efficient because there could be hundreds of words in the context within only a few sentences. Also, our attention is static in that only the nearest two sentences around the decision point search for relevant information; it differs from dynamic attention~\cite{bahdanau2014neural}, which buries more important sentences under less important context.

Our model was evaluated on transcripts of nearly 3,000 episodes of TV talk shows. 
In our experiments, modern neural networks consistently outperform traditional methods that use handcrafted features. Ablation tests confirm the effectiveness of context; the proposed hierarchical RNN with sentence-level static attention can better utilize such contextual information, and significantly outperforms non-attention neural networks.
The results show that our tailored model is especially suited to the task of text-based speaker change detection.

\section{Related Work}

Traditional speaker change detection (SCD) deals with audio input  and is a key step for  \textit{speaker diarization} (determining ``who spoke when?'')~\cite{Anguera2012Speaker}. A typical approach is to compare consecutive sliding windows of input with spectral features, pitch, or silence points~\cite{kemp2000strategies,lu2005unsupervised}.  Our paper differs from the above work and focuses on textual input, which is useful in various scenarios, for example, processing dialog transcripts where speaker identities are missing (e.g., OpenSubtitle)~\cite{li2015diversity}, and enhancing audio SCD with textual information~\cite{li2009improving}.

Nowadays, text-based dialog analysis has been increasingly important, as
surface acoustic features are insufficient for semantic understanding in conversations.
Previous research has addressed a variety of tasks, ranging from dialog act classification~\cite{Shen+2016} to user intent modeling~\cite{userintent}. In our previous study, we address the problem of session segmentation in text-based human-computer conversations~\cite{Song+2016}. Without enough annotated data, we apply a heuristic matching approach, thus the task being unsupervised.
Li et al.~\cite{li2009improving} enhance audio-based SCD  with transcribed text, and they are also in the unsupervised regime. 
By contrast, this paper adopts a supervised setting as we have obtained massive, high-quality labels of speaker identities from the Cable News Network website.

As described in Section~\ref{sec:introduction}, we formulate our task as a sentence-pair classification problem.
Previous studies have utilized convolutional/recurrent neural networks (CNNs/RNNs) to detect the relationship (e.g., paraphrase and logical entailment) between two sentences~\cite{bowman}; Rockt\"aschel {et al.}~\cite{rocktaschel2015reasoning} equip RNN with attention mechanisms. These studies do not consider contextual information.
In our scenario, the context appears on both sides of the decision point, and we carefully design the neural architecture to better use such contextual information.

\section{Approach}

In this section, we describe the proposed approach in detail. Figure~\ref{fig:model} shows the overall architecture of our model, which has three main components: a sentence encoder, a context encoder, and an attention-based matching mechanism.

\subsection{Sentence Encoder}\label{app:sentence_encoder}

We use a recurrent neural network (RNN) with long short term memory (LSTM) units to encode a sentence as a vector (also known as a \textit{sentence embedding}), shown in Figure~\ref{fig:model}a.

An RNN is suited for processing sequential data (e.g., a sentence consisting of several words) as it keeps a hidden state, changing at each time step based on its previous state and the current input.
But vanilla RNNs with perceptron-like hidden states suffer from the problem of \textit{vanishing or exploding gradients}, being less effective to model long dependencies. LSTM units alleviate the problem by better balancing input and its previous state with gating mechanisms. For convenience, we use LSTM's  final state (corresponding to the last word in a sentence) as the sentence embedding.

Formally, let $\bm x^{(t)}$ be the embedding of the $t$-th word in a sentence, and $\bm h^{(t-1)}$ be the last step's hidden state. We have
\begin{align}
[\bm i^{(t)}; \bm f^{(t)}; \bm o^{(t)}] &= \sigma( W  \bm x^{(t)}  +  U\bm h^{(t-1)} + \bm b) \\
\bm g^{(t)} &= \tanh( W_g  \bm x^{(t)} +  U_g \bm h^{(t-1)} + \bm b_g) \\
\bm c^{(t)}  &= \bm i^{(t)}\otimes \bm g^{(t)}  + \bm c^{(t-1)}\otimes \bm f^{(t)}\\
\bm h^{(t)}  &= \bm o^{(t)} \otimes \tanh( \bm c^{(t)} )
\end{align}
where $W$'s and $U$'s are weights, and $\bm b$'s are bias terms. $\otimes$ denotes element-wise product, $\sigma$ the sigmoid function. $\bm i^{(t)}$, $\bm f^{(t)}$, and $\bm o^{(t)}$ are known as gates, and $\bm h^{(t)}$ is the current step's hidden state.

\subsection{Context Encoder}\label{app:context_encoder}

Another LSTM-RNN encodes contextual information over sentence vectors, shown in Figure~\ref{fig:model}b. 
Since we model our task as a matching problem, we apply the RNN to the decision point's both sides separately, the resulting vectors of which are concatenated as an input of prediction.

It should be noticed that our LSTM-RNN goes from faraway sentences to the nearest ones (called \textit{critical sentences}) on both sides of the current decision point. We observe that nearer sentences play a more important role for prediction; that an RNN is better at keeping recent input information by its nature. Hence, our treatment is appropriate.

Besides, our neural network is hierarchical in that it composites sentences with words, and discourses with sentences. It is similar to the hierarchical autoencoder in~\cite{li2015hierarchical}. Other studies apply a single RNN over a discourse, also achieving high performance in tasks like machine comprehension~\cite{seo2016bidirectional}. In our scenario, however, the sentences (either context or critical ones) are not necessarily uttered by a single speaker. Experiments in Section~\ref{exp:performance} show that hierarchical models are more suitable for SCD.

\subsection{Our Attention Mechanism}\label{app:attention}

We use an attention mechanism to better utilize contextual information. Attention-based neural networks are first proposed to dynamically focus on relevant words of the source sentence in machine translation~\cite{bahdanau2014neural}. In our scenario, we would like to match a critical sentence with all utterances on the other side of the decision point (Figure~\ref{fig:model}c). That is to say, the attention mechanism is applied to the sentence level, different from other work that uses word-level attention. Our method is substantially more efficient because a context of several utterances could contain hundreds of words.

Considering $t$ sentences (context size being $t-1$) before and after the decision point, respectively, we have $2t$ sentences in total, namely
$\bm s_p^{(1)}\!\rightarrow\!\bm s_p^{(2)}\!\rightarrow\!\cdots\!\rightarrow\!\bm s_p^{(t)}\quad\quad \bm s_f^{(t)}\!\leftarrow\!\cdots\!\leftarrow\! \bm s_f^{(2)}\!\leftarrow\!\bm s_f^{(1)}$, where subscripts $p$ and $f$ refer to previous and future utterances around the current decision point; the arrows indicate RNN's directions. 

In our attention mechanism, a critical sentence, e.g., $\bm s_p^{(t)}$, focuses on all sentences on the other side of the decision point $\bm s_f^{(1)}, \cdots, \bm s_f^{(t)}$, and aggregates information weighted by a probabilistic distribution $\bm \alpha_p\in\mathbb{R}^t$, i.e.,
\begin{align}\label{eqn:att1}
\tilde{\alpha}_p^{(i)}  &= \bm u_a^\top\tanh\big(W_a \big[\bm s_p^{(t)}; \bm s_f^{(i)}\big]\big)\\\label{eqn:att2}
\alpha_p^{(i)} &= \operatorname{softmax}\left(\tilde\alpha_p^{(i)}\right) = \frac{\exp\big\{\tilde{\alpha}_p^{(i)}\big\}}{\sum_{j=1}^t\exp\big\{\tilde{\alpha}_p^{(j)}\big\}}
\end{align}
Here, $\bm s_p^{(t)}$ is concatenated with $\bm s_f^{(i)}$, processed by a two-layer perceptron (with parameters $W_a$ and $\bm u_a$).
$\tilde \alpha_i$ is a real-valued measure, normalized by $\operatorname{softmax}$ to give the probability $\alpha_i$.
The aggregated information, known as an \textit{attention vector}, is 
\begin{align}
\bm m_p &= \sum_{i=1}^t \alpha_p^{(i)} \cdot\bm s_f^{(i)}\label{eqn:att3}
\end{align}

Likewise,  the other critical sentence $\bm s_f^{(t)}$ yields an attention vector $\bm m_f$. They are concatenated along with LSTM's output of the critical sentences for prediction. In other words, the input of $\operatorname{softmax}$ is $[\bm s_p^{(t)}; \bm s_f^{(t)};\bm m_p; \bm m_f]$.

It should be pointed out that, our attention is static, as only critical sentences search for relevant information using Equations~(\ref{eqn:att1})--(\ref{eqn:att3}). It resembles a variant in~\cite{rocktaschel2015reasoning}, but differs from common attention where two LSTMs interact dynamically along their propagations (Figure~\ref{fig:model}d). Such approach may bury the critical sentences; it leads to slight performance degradation in our experiment, as we shall see in Section~\ref{exp:performance}. 

\section{Experiments}\label{sec:exp}


\subsection{Dataset Collection}\label{exp:dataset}

We crawled transcripts of nearly 3,000 TV talk shows from the Cable News Network (CNN) website,\footnote{\url{http://transcripts.cnn.com} (\textit{CNN} here should not be confused with a convolutional neural network.)} and extracted main contents from the original html files. The transcripts contain speaker identities, with which we induced speaker changes in the dialog.

The crawled dataset comprises 1.5M utterances. We split training, validation, and test sets by episodes (TV shows) at a ratio of 8:1:1. In other words, each episode appears in either the training set, or the val/test sets.
This prevents utterance overlapping between training and prediction, and thus is a more realistic setting than splitting by utterances. 

We notice that, our corpus is larger than previous ones by magnitudes: the 1997 HUB4 dataset, for example, is 97 hours long, whereas our TV shows are estimated to be 3,000 hours (each episode roughly lasting for an hour),
which are more suitable for training deep neural networks.

In the dataset, speaker changes count to 25\%. We thus used $F_1$-measure in addition to accuracy as metrics, i.e., $F_1=2P\cdot R/(P+R)$, where $P=\frac{\text{\#correctly detected changes}}{\text{\#detected changes}}$ is the \textit{precision}, and $R=\frac{\text{\#correctly detected changes}}{\text{\#all changes}}$ is the \textit{recall}.

	\begin{table}[!t]
		\centering
		\resizebox{\linewidth}{!}{
			\begin{tabular}{|l|c|c|c|c|}
				\hline
				\!\!Model & Acc. & $F_1$ & $P$ & $R$ \\
				\hline
				\hline
				\!\!Random guess & 61.8 & 25.4 & 26.0 & 25.0 \\
				\!\!Logistic regression w/ (uni+bi)-gram\!\!\! &80.5 & 50.9 & 73.0  &  39.0  \\
				\!\!DNN w/ (uni+bi)-gram  &  76.6   & 56.5  &  54.4 & 58.8 \\
				\hline
				\hline
				\!\!CNN w/o context            & 77.8  & 57.8  & 56.8  & 58.9  \\
				\!\!RNN w/o context            & 83.3  & 63.9 & 72.5 & 57.1 \\
				\hline
				\!\!RNN w/ context (non-hierarchical) \!\!\!\!\! &  83.7 &  65.7 &72.6 &  60.0  \\
				\!\!RNN w/ context (hierarchical) &85.1  & 69.2 & 74.6 & 64.6 \\
				\quad\quad\quad\quad\quad\ \ + static attention  & \textbf{89.2}  & \textbf{78.4} & \textbf{81.5} & \textbf{75.6} \\
				\hline
			\end{tabular}}
			\caption{Model performance (in \%). Here, LSTM units are used in RNN, but omitted in the table for brevity. The context size is 8, chosen by validation (deferred to Figure~\ref{fig:analysis}).}
			\vspace{-.7cm}
			\label{tab:performance}
		\end{table}
		
		\subsection{Settings}
		
		We set all neural layers, including word embeddings, to 200 dimensional. Since our dataset is large, we randomly initialized word embeddings, which were tuned during training. 
		We used the Adam optimizer with mini-batch update (batch size being 100).
		Other hyperparameters were chosen by validation: dropout rate from $\{0.1, 0.3\}$ and initial learning rate from $\{3\times10^{-4},  9\times10^{-4}\}$. 
		
		We had several baselines with handcrafted features: we extracted unigram and bigram features of the critical sentences as two vectors, which are concatenated for prediction. We applied logistic regression and a 3-layer deep neural network (DNN) as the classifier; the former is a linear model whereas the latter is nonlinear. DNN's hidden dimension was set to 200, which is the same as our neural network. 
		
		A convolutional neural network (CNN) is also included for comparison. It adopts a window size of 3, and a max-pooling layer aggregates extracted features.
		All competing neural models (including DNN and CNN) were tuned in the same manner, so our comparison is fair.
		
		\subsection{Performance}\label{exp:performance}
		
		Table~\ref{tab:performance} presents the performance of our model as well as baselines. 
		As shown, all modern neural networks (CNNs/RNNs with word embeddings) are consistently better than methods using handcrafted features of unigrams and bigrams. 
		Because we have applied a 3-layer DNN to these features, we believe the performance improvement is not merely caused by using a better classifier, but the automatic
		feature/representation learning nature of modern neural networks.
		
		For neural network-based sentence encoders, we compared LSTM-RNN with CNN.
		Results show that LSTM-RNN outperforms CNN by 6\% in terms of both accuracy and $F_1$-measure.

		To cope with context, the simplest approach, perhaps, is to use an RNN to go through surrounding utterances of the critical sentences, denoted as \textit{non-hierarchical} in Table~\ref{tab:performance}. Using contextual information yields an $F_1$ improvement of 2\%. This controlled experiment validates the usefulness of context for SCD.
		The hierarchical RNN introduced in Sections~\ref{app:sentence_encoder} and~\ref{app:context_encoder} further improves the $F_1$-measure by  3\%.
		With sentence-level static attention, our model achieves the highest performance of 89.2\% accuracy and 78.4\% $F_1$-measure.
		
\begin{figure}[!t]
			\centering

			\includegraphics[width=.7\linewidth]{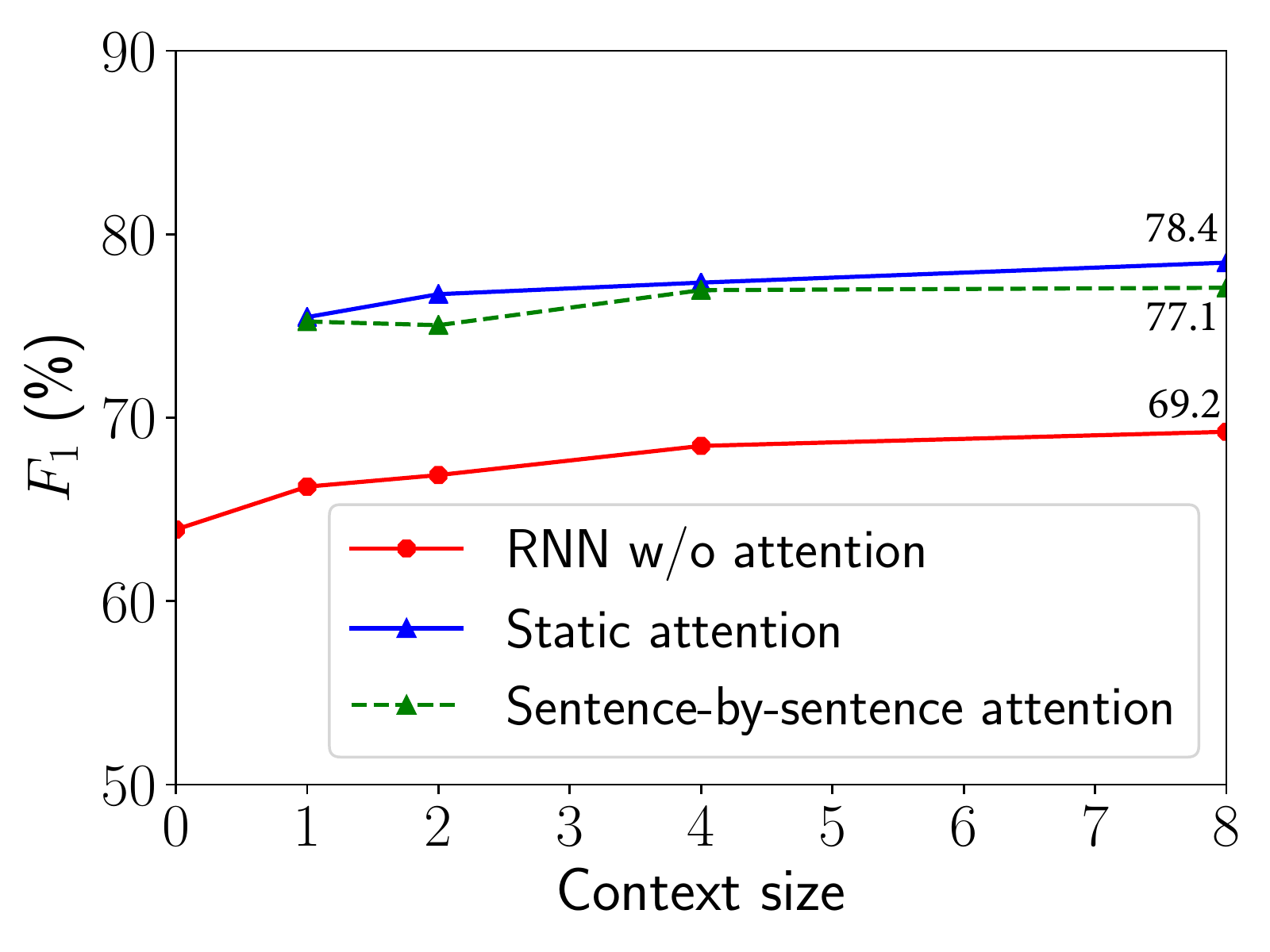}
			\caption{Effect of the context size and different attention mechanisms. The context size is the number of utterances, excluding the critical sentence, on each side of the decision point;
				a size of 0 refers to no context.}
			\label{fig:analysis}

		\end{figure}

		We would like to have in-depth analysis regarding how the context size and different attention mechanisms affect our model.
		The context size was chosen by validation from $\{1, 2, 4,8\}$.\footnote{Due to efficiency concerns, we did not try larger context sizes.} As shown in Figure~\ref{fig:analysis}, even a single context sentence (on each side of the decision point) improves the performance by 2\%; with more surrounding utterances, the performance grows gradually. 
		Moreover, attention-based neural networks significantly outperform non-attention models by a margin of 10\%. 
		We also tried a dynamic sentence-by-sentence attention mechanism, similar to most existing work~\cite{bahdanau2014neural}. 
		As analyzed in Section~\ref{app:attention}, such model buries critical sentences and thus slightly hurts the performance by 1--2\% $F_1$-measure (green dashed line in Figure~\ref{fig:analysis}).
		The experiments verify the effectiveness of our hierarchical RNN with sentence-level, static attention.
		
		\noindent\textbf{Case study.} Table~\ref{tab:case_study} showcases a dialog snippet. 
		In the example, our model makes an error as it fails to detect the change between Sentences 4--5.
		However, it is even hard for humans to judge this particular change, because the word \textit{absolutely} also goes fluently into the next sentence.
		For other utterances with more substance, the neural network correctly joins Sentences 1--2 and 5--6, as well as segments Sentences 2--3 and 3--4,
		showing that our proposed model can effectively capture the semantics of these sentences.
		
		\begin{table}[!t]
			\centering

			\resizebox{0.93\linewidth}{!}{
				\begin{tabular}{c|p{.53\linewidth}|ccc}
					\toprule
					\multirow{2}{*}{\textbf{ID}}&\multicolumn{1}{c|}{\multirow{2}{*}{\textbf{Utterances}}} & \multicolumn{2}{c}{\textbf{Speaker Changes?}} & \multirow{2}{*}{\textbf{Correct?}}\\
					&                 & Predicted & Truth & \\
					\midrule
					\multirow{2}{*}{1}&\!  {\color{Crimson}there 's no question the deficit halts on both sides of the aisles . } & \multirow{5}{*}{No} & \multirow{5}{*}{No} & \multirow{5}{*}{\cmark}\\
					\cmidrule{1-2}
					\multirow{3}{*}{2}&\! \color{Crimson} cbo , wall street , everyone will have a say into this , including workers and future retirees . &  \multirow{6}{*}{Yes} & \multirow{6}{*}{Yes} & \multirow{6}{*}{\cmark}\\
					\cmidrule{1-2}
					\multirow{2}{*}{3}&\! \color{DodgerBlue}and your saying it 's both\ \hspace{1cm}\color{DodgerBlue} parties ? &  \multirow{4}{*}{Yes} & \multirow{4}{*}{Yes} & \multirow{4}{*}{\cmark}\\
					\cmidrule{1-2}
					4&\ \hspace{-1.2mm}\color{Crimson}absolutely . &  \multirow{3}{*}{No} & \multirow{3}{*}{Yes} & \multirow{3}{*}{\xmark}\\
					\cmidrule{1-2}
					\multirow{2}{*}{5}&\! \color{Peru}the key , though , is the first six months . &  \multirow{5}{*}{No} & \multirow{5}{*}{No} & \multirow{5}{*}{\cmark}\\
					\cmidrule{1-2}
					\multirow{3}{*}{6}&\! \color{Peru}you say it 's not going to get through in these first couple months . \\
					\bottomrule
				\end{tabular}}
				\caption{Case study. Colors indicate speaker identities (they are used to infer groundtruth speaker changes, but cannot be seen during prediction).}\label{tab:case_study}
				\vspace{-.3cm}
			\end{table}

			\section{Conclusion}
			
			In this paper, we proposed a static sentence-level attention LSTM-RNN for text-based speaker change detection.
			Our model uses an LSTM-RNN to encode each utterance into a vector, based on which another LSTM-RNN integrates contextual information, before and after a particular decision point, respectively. A static sentence-level attention mechanism is also applied to enhance information interaction.
			We crawled dialog transcripts from Cable News Network TV talk shows for evaluation. Experimental results demonstrate the effectiveness of our approach. In particular, in-depth analysis validates that contextual information is indeed helpful for speaker change detection, and that our tailored model can make better use of context than other neural networks.
			
			
			\bibliographystyle{ACM-Reference-Format}
			\bibliography{mybib2}

\end{document}